\documentclass[journal,twoside,web]{ieeecolor}

\usepackage{lcsys}
\usepackage{cite}
\usepackage{amsmath,amssymb,amsfonts}
\usepackage{algorithmic}
\usepackage{graphicx}
\usepackage{textcomp}
\usepackage{amsmath}
\usepackage{amsfonts}
\usepackage{amssymb}
\usepackage{mathrsfs}

\newtheorem{theorem}{Theorem}
\newtheorem{corollary}[theorem]{Corollary}
\newtheorem{lemma}[theorem]{Lemma}

\begin{document}

\title{Dynamic Regret of Online Mirror Descent for Relatively Smooth Convex Cost Functions}

\author{Nima Eshraghi, \IEEEmembership{Student member, IEEE}, and Ben Liang, \IEEEmembership{Fellow, IEEE} 
\thanks{N. Eshraghi and B. Liang are with the Department of Electrical and Computer Engineering, University of Toronto, Ontario, Canada. ({\tt\small e-mail: \{neshraghi, liang\}@ece.utoronto.ca}). \\
This work was funded in part by the Natural Sciences and Engineering Research Council of Canada.}
}

\maketitle
\thispagestyle{empty}

\begin{abstract}
The performance of online convex optimization algorithms in a dynamic environment is often expressed in terms of the dynamic regret, which measures the decision maker's performance against a sequence of time-varying comparators. In the analysis of the dynamic regret, prior works often assume Lipschitz continuity or uniform smoothness of the cost functions. 
However, there are many important cost functions in practice that do not satisfy these conditions. In such cases, prior analyses are not applicable and fail to guarantee the optimization performance.
In this letter, we show that it is possible to bound the dynamic regret, even when neither Lipschitz continuity nor uniform smoothness is present. We adopt the notion of relative smoothness with respect to some user-defined regularization function, which is a much milder requirement on the cost functions.
 We first show that under relative smoothness, the dynamic regret has an upper bound based on the path length and functional variation. We then show that with an additional condition of relatively strong convexity, the dynamic regret can be bounded by the path length and gradient variation.
These regret bounds provide performance guarantees to a wide variety of online optimization problems that arise in different application domains. Finally, we present numerical experiments that demonstrate the advantage of adopting a regularization function  under which the cost functions are relatively smooth. 
\end{abstract}


\section{Introduction}
\label{sec:intro}

\IEEEPARstart{W}{e study} the problem of online convex optimization, which can be modeled by a sequential decision-making process over a finite number of rounds $T$. In every round $t \in \{1, 2, \ldots, T\}$, the decision maker chooses a point $x_t$ from a convex set $\mathcal{X}$ based on the information from previous rounds. Then, the convex cost function $f_t(x): \mathcal{X} \rightarrow \mathbb{R}$ is revealed to the decision maker, which suffers the corresponding cost $f_t(x_t)$. The goal of the decision maker is to minimize some regret, which is the difference between the cumulative cost of the decision maker and that of an optimal fixed point chosen in hindsight with knowledge of $f_t(x)$ for all rounds. Most early works consider the static regret:
\begin{align}\label{def_st_reg}
\mathrm{Reg}_T^\mathrm{s} = \sum_{t=1}^{T} f_t(x_t) - \underset{x \in \mathcal{X}}{\text{argmin}} \sum_{t=1}^{T} f_t(x).
\end{align}
The benchmark variable in \eqref{def_st_reg} is a \textit{static} point that resides in the feasible set $\mathcal{X}$. In the literature of online learning and control, there are many algorithms that guarantee sublinear upper bounds on the static regret under various settings \cite{poor21, sparsetv21, zinkevich2003online, shalev2012online}. 

Such a static metric can accurately reflect the decision maker's performance as long as the static benchmark performs consistently well over different rounds. However, this may not always hold in a dynamic environment, where the sequence of cost functions are time-varying. Thus, the \textit{dynamic regret} has been proposed as a more stringent metric that measures the algorithm performance against a time-varying sequence. Previous studies often focus on the the sequence of minimizers in their regret analysis \cite{antoine21, eshraghi2020distributed, yangb16, mokhtari2016online, zhang2017improved, chang2020unconstrained}.
In this work, we target a more general form of the dynamic regret that allows comparison against an arbitrary comparator sequence $\{u_1, u_2, \ldots, u_T\}$\cite{zinkevich2003online}:
\begin{align}\label{def_gen_dyn_reg}
\mathrm{Reg}_T^\mathrm{d} = \sum_{t=1}^{T} f_t(x_t) - \sum_{t=1}^{T} f_t(u_t).
\end{align}

Since the online cost functions can fluctuate arbitrarily, obtaining a sublinear upper bound on the dynamic regret may not be possible. Therefore, the dynamic regret is usually expressed in terms of certain \textit{regularity measures} of the comparators or the cost function sequence. The regularity measures reflect how fast an environment evolves as time progresses. Previous works \cite{zinkevich2003online,   yangb16, hall2013dynamical, hall2015, zhao21proximal, cumul21Yi, eshraghi2022acc, zhang2017improved,  besbes2015non, jadbabaie2015online, mokhtari2016online,  chang2020unconstrained} have utilized different regularity measures to bound the dynamic regret, namely the path length, gradient variation, and functional variation. The path length of an arbitrary sequence $\{u_1, u_2, \ldots, u_T\}$ is defined~as 
\begin{align} \label{eq_path_length}
  C_T = \sum_{t=1}^{T} \| u_{t+1} - u_t\|,  
\end{align}
which measures the variation in the comparator sequence, where $\|\cdot\|$ could be any norm. Several online learning algorithms provide an $O(\sqrt{T} (1+ C_T))$ upper bound on the dynamic regret of convex cost functions \cite{zinkevich2003online, hall2013dynamical, hall2015}, which can be improved to $O(\sqrt{T C_T})$ when prior knowledge of $C_T$ and $T$ is available \cite{zhao21proximal}. The path length has also been recently used in the study of online convex optimization with constraint violation \cite{cumul21Yi}, where upper bounds of $O(\sqrt{T(1+C_T)})$ and $O(\sqrt{T})$ are derived on the dynamic regret and cumulative constraint violation, respectively. Furthermore, the dynamic regret can also be bounded by both the path length and its squared form using multiple mirror descent steps per online round \cite{eshraghi2022acc}. In addition to the path length, gradient variation is another regularity measure often utilized in the online learning literature \cite{jadbabaie2015online, rakhlin2013optimization, chiang12}. It reflects how fast the gradient of the online functions changes. Here, we use a non-squared version of the gradient variation, which is defined as 
\begin{align}\label{eq_gradient_var_2}
G_T = \sum_{t=1}^{T} \sup_{x \in \mathcal{X}}\|\nabla f_t(x) - \nabla f_{t-1}(x)\|_*,
\end{align}
where $\|x\|_* = \sup_y \{x^Ty  | \|y\| \leq 1\}$ represents the dual norm. Another regularity measure related to the sequence of online functions is functional variation \cite{chang2020unconstrained, jadbabaie2015online, besbes2015non}, defined as
\begin{align}\label{eq_functional_var}
    V_T = \sum_{t=1}^{T} \sup_{x \in \mathcal{X}} | f_{t+1}(x) - f_t(x) |,
\end{align}
which collects the variation in the cost functions over time.

In the analysis of dynamic regret, prior works often impose assumptions such as Lipschitz continuity and/or uniform smoothness \cite{zinkevich2003online,  yangb16, zhao21proximal, cumul21Yi, hall2013dynamical, hall2015, zhang2017improved, jadbabaie2015online, mokhtari2016online, chang2020unconstrained, besbes2015non}. However, some cost functions that arise in well-known applications do not satisfy these conditions, e.g., the Poisson inverse problem, the D-optimal design problem, and support vector machine training. 
To address this limitation, a generalization of these traditional assumptions is required. In this work, we remove the restrictive assumptions of Lipschitz continuity and uniform smoothness.

Instead, we adopt the notion of relative smoothness, which allows measuring the cost functions behavior relative to a user-specific function \cite{lu2018relatively}. Relative smoothness generalizes the traditional form of smoothness used in earlier analyses and provides more flexibility in functions. Consequently, it is applicable to a broader class of cost functions. Such generalization has been proposed recently in the study of \textit{offline} optimization methods based on mirror descent \cite{bauschke2017descent, lu2018relatively, lu2019relative, bauschke2019linear}. However, this is not applicable to many systems in practice since they are often time-varying, requiring an online solution. Performance analysis in the online setting is more challenging due to the appearance of some dynamic terms that require careful handling to reflect the speed of the changes in the problem environment. There is no prior work to study the feasibility of generalization to relative smoothness in the online setting.\footnote{In the online setting,  the recent work of \cite{zhou2020regret} studies the static regret without Lipschitz continuity. They introduce the notion of Riemann-Lipschitz continuity, to bound the static regret of online mirror descent under this assumption. However, as discussed above, the dynamic regret often is a more suitable measure of performance.}

 As far as we are aware, this is the first study to bound the dynamic regret of online mirror descent for relatively smooth convex functions. First, we show that the dynamic regret has an upper bound of $O(1 + C_T + V_T)$. This compares favorably to the closest related works in \cite{ hall2013dynamical, hall2015}, which obtain the upper bound of $O(\sqrt{T} (1 + C_T))$ for online mirror descent under the Lipschitz continuity assumption. Besides removing the requirement of Lipschitz continuity, our bound further reduces the dependence on $T$ and can be much smaller especially when the problem environment does not drift too fast. Second, we show that when the cost functions are in addition  relatively strongly convex, the dynamic regret can be further tightened to $O(1+C_T + \min(V_T, G_T))$. Thus, our results show that even when the cost functions are not Lipschitz continuous or uniformly smooth, it is still possible to guarantee performance in terms of the dynamic regret, by leveraging relative smoothness. Finally, our numerical experiments demonstrate substantial improvement in the performance of online mirror descent when the regularization function is chosen to provide relative smoothness.

\section{Dynamic Regret of Online Mirror Descent}
\label{sec_regret}

 We consider the standard problem of online optimization with respect to a sequence of convex cost functions over a finite number of rounds, denoted by $T$. At the beginning of every round $t$, the decision maker submits a decision represented by $x_t$, which is taken from a convex and compact set $\mathcal{X}$. Then, the cost function of the current round $f_t(\cdot)$ is revealed,  and the decision maker becomes aware that it has suffered the corresponding cost $f_t(x_t)$. The decision maker then updates its decision in the next round.

 The mirror descent algorithm is a classical method for convex optimization problems.  An appealing feature of mirror descent is the extension of the projection step using the Bregman divergence instead of the Euclidean distance, which makes the algorithm applicable to a wide range of problems. The Bregman divergence with respect to the function $r(\cdot)$ is defined as 
\begin{align} \nonumber
D_{r}(x,y) = r(x) - r(y) - \langle \nabla r(y), x - y \rangle.
\end{align}
The Bregman divergence is a general way of measuring the distance through the lens of function $r(\cdot)$. An instructive example is the Bregman divergence associated with the squared $\ell_2$-norm, i.e.,  $r(x) = \frac{1}{2}\|x\|_2^2$. In this case, the Bregman divergence reduces to the Euclidean distance. In another example, the Bregman divergence corresponding to the choice of $r(x) = \sum_{i=1}^{d} x_i \log(x_i) - x_i$ on the $d$-dimensional probability simplex recovers the Kullback-Leibler divergence. Many other interesting examples are given in \cite{bauschke2019linear}.

The online version of  mirror descent \cite{shalev2012online} is given by
\begin{align} \label{def_omd}
    x_{t+1}= \underset{x \in \mathcal{X}}{\text{argmin}}\Big\{\langle \nabla f_t(x_t), x \rangle + \frac{1}{\eta_t} D_r(x, x_t) \Big\} 
\end{align}
where $\eta_t$ is the step size, and $D_{r}(x, y)$ is the Bregman divergence corresponding to the function $r(\cdot)$. In the context of mirror descent, $r(\cdot)$ is often called the \textit{regularization function}.  The term with the Bregman divergence helps to limit the changes in the sequence of decisions from one round to the next. In particular, the update in \eqref{def_omd} suggests that the decision maker aims to stay close to the current decision $x_{t}$ as measured by the Bregman divergence, while taking a step in a direction close to the negative gradient to reduce the current cost at round $t$.
We make the following assumption, which is common in the study of online mirror descent \cite{jadbabaie2015online, shahrampour2017distributed}:

\textit{Assumption 1.} The Bregman divergence satisfies a Lipschitz condition of the form
 \begin{align} \label{eq_bregman_lip}
|D_{r}(x, z) - D_{r}(y, z)| \leq \gamma \|x - y\|,~ \forall x,y,z \in \mathcal{X},
\end{align}
where $\gamma$ is a positive constant. 

\noindent Note that when the function $r(\cdot)$ is Lipschitz continuous on the feasibility domain, the Lipschitz condition on the Bregman divergence is automatically satisfied.
 
 In this work, we do not require the cost functions to be Lipschitz continuous while the condition stated in \eqref{eq_bregman_lip}
serves as its replacement.  In online optimization, the cost functions are revealed to the decision maker sequentially over time, and we have no control over them. In contrast, we have control over the choice of the regularization function, which can benefit from a careful design to allow properties such as Lipschitz continuity. 
 Therefore, the condition in \eqref{eq_bregman_lip} is indeed a milder requirement compared with the assumption of Lipschitz continuity of the cost functions. 

\subsection{Relatively Smooth Cost Functions}\label{sec_rel_smooth}

In this section, we consider relatively smooth cost functions that may not be Lipschitz continuous or uniformly smooth. The notion of relative smoothness is proposed in \cite{lu2018relatively}, which measures the smoothness relative to a user-specific function. Thus, it does not require the specification of any norm. We provide below the definitions of uniform smoothness and relative smoothness.

\textit{Definition 1:}\label{def_smooth} A function $f : \mathcal{X} \rightarrow \mathbb{R}$ is uniformly smooth with modulus $\beta$, i.e., $\beta$-smooth,  with respect to some norm $\|\cdot\|$, if there exists a positive constant $\beta$ such that 
\begin{align} \nonumber
f(y) \leq f(x) + \langle \nabla f(x), y - x \rangle + \frac{\beta}{2} \| y - x \|^2, ~\forall x,y \in \mathcal{X}.
\end{align}
An equivalent definition is that $\nabla f(\cdot)$ is Lipschitz continuous, i.e., $\|\nabla f(x) - \nabla f(y)\| \leq  \beta \|x - y\|$,  $ \forall  x, y \in \mathcal{X}$.
\color{black}

\textit{Definition 2:} \label{def_rel_smooth} A function $f : \mathcal{X} \rightarrow \mathbb{R}$ is $\beta$-smooth relative to $r(\cdot)$, if there exists a positive constant $\beta$ such that
\begin{align} \nonumber
f(y) \leq f(x) + \langle \nabla f(x), y - x \rangle + \beta D_r(y,x), ~\forall x,y \in \mathcal{X}.
\end{align}
 
 We note that the relative smoothness in Definition~2 generalizes the uniform smoothness in Definition~1, which is commonly assumed in the literature of online convex optimization \cite{ zhang2017improved, chang2020unconstrained, chiang12, zhao2020dynamic}. In Definition~2, the Bregman divergence naturally serves as a distance measure. It replaces the norm squared in Definition 1. Therefore, the smoothness of $f(\cdot)$ does not depend on any norm and is instead measured with respect to the function $r(\cdot)$.   In particular, by setting $r(x) = \frac{1}{2}\|x\|^2$, relative smoothness specializes to the uniform smoothness. Furthermore, from  \textit{Proposition}~1.1 in \cite{lu2018relatively}, an equivalent form of Definition~2 is 
\begin{align}\label{def_rel_smooth_eq}
 \nabla^2 f(x) \preceq \beta~\nabla^2 r(x),  ~\forall x \in \mathcal{X},
\end{align}
which establishes a simple condition on the Hessian matrices of the two functions.

We make the following assumption for the analysis in this subsection.

\textit{Assumption 2.} \label{assumption_relative}
 The cost functions $f_t(\cdot)$ are convex and $\beta$-smooth relative to the regularization function $r(\cdot)$.

As an example, the D-optimal design problem \cite{lu2018relatively} satisfies the condition stated in Assumption~2, but it is neither Lipschitz continuous nor uniformly smooth.  There are many other important relatively smooth functions that arise in various application domains, such as the Poisson inverse problem \cite{bauschke2017descent} and minimum-volume covering ellipsoid \cite{lu2018relatively, bauschke2019linear}. A systematic way of choosing a proper function $r(\cdot)$ is presented in \cite{lu2018relatively} for any objective function whose norm of subgradients are bounded by a polynomial in either $\sum_{i=1}^d \frac{1}{x_i}$ or $\|x\|_2$. It is a useful construction to reveal the relative smoothness of a wide range of cost functions.

We are now ready to upper bound the dynamic regret of online mirror descent under relative smoothness.

\begin{theorem} \label{thm_reg_rel_smooth}
Under Assumptions 1 and 2, the dynamic regret of online mirror descent with fixed step size $\eta = 1/\beta$ satisfies 
\begin{align}
 \sum_{t=1}^{T}  \Big(f_t( &  x_t)  - f_t(u_t)\Big) \leq    \nonumber\\
& ~\beta R+ f_1(x_1) - f_{T+1}(x_{T+1}) + \gamma\beta C_T +  V_T, \nonumber
\end{align}
for any feasible sequence $\{u_1, u_2, \ldots, u_T\}$, where  $R = \max_{x,y \in \mathcal{X}} D_r(x,y)$,  $\gamma$ is  the Lipschitz constant associated with the Bregman divergence, and $C_T$ and $V_T$ denote the path length and functional variation, as defined in \eqref{eq_path_length} and \eqref{eq_functional_var}, respectively. 
\end{theorem}

The proof of Theorem~\ref{thm_reg_rel_smooth} is given in App.~\ref{sec_proof_thm_rel_smooth}.

\textit{Remark 1.}
 The dynamic regret of standard online mirror descent under uniform smoothness has not been studied in prior works. However, the works  \cite{hall2013dynamical, hall2015}  have shown that for convex and Lipschitz continuous cost functions, the dynamic regret of online mirror descent is bounded by $O(\sqrt{T}(1+C_T))$. 
 Theorem \ref{thm_reg_rel_smooth} shows that even when the Lipschitz continuity requirement is replaced by relative smoothness, the dynamic regret can still be upper bounded. Furthermore, the new $O(1 + C_T + V_T)$ bound removes the dependency on $\sqrt{T}$ and relates only to the regularity measures $C_T$ and $V_T$. 

\subsection{Relatively Smooth and Strongly Convex Cost Functions}\label{sec_rel_smooth_str_cvx}

In this part, we consider cost functions that are relatively smooth and strongly convex with respect to the same function $r(\cdot)$. We provide the formal definitions of uniform strong convexity and relative strong convexity below:

\textit{Definition 3:}\label{def_strcvx} A function $f : \mathcal{X} \rightarrow \mathbb{R}$ is uniformly strongly convex with modulus $\lambda$, i.e., $\lambda$-strongly convex,  with respect to some norm $\|\cdot\|$, if there exists a positive constant $\lambda$ such that 
\begin{align} \nonumber
f(x) + \langle \nabla f(x), y - x \rangle + \frac{\lambda}{2} \| y - x \|^2 \leq f(y), ~\forall x,y \in \mathcal{X}.
\end{align}

\textit{Definition 4:} \label{def_rel_strcvx} A function $f : \mathcal{X} \rightarrow \mathbb{R}$ is $\lambda$-strongly convex relative to $r(\cdot)$, if there exists a positive constant $\lambda$ such that
\begin{align} \nonumber
f(x) + \langle \nabla f(x), y - x \rangle + \lambda D_r(y,x) \leq f(y), ~\forall x,y \in \mathcal{X}.
\end{align}

We make the following assumption in this subsection.

\textit{Assumption 3.}
The cost functions $f_t(\cdot)$ are  $\beta$-smooth and $\lambda$-strongly convex, both relative to the regularization function $r(\cdot)$.
Furthermore,  $r(\cdot)$ is $1$-strongly convex with respect to some norm $\|\cdot\|$.

We note that the conditions stated in Assumption 3 imply $\lambda \leq \beta$. Furthermore, since $r(\cdot)$ is strongly convex with respect to a norm $\|\cdot\|$, Assumption 3 implies the strong convexity of $f_t(\cdot)$ with respect to the same norm. We also note that the strong convexity of the regularization function is a standard assumption, commonly used in the analysis of online mirror descent \cite{shalev2012online,  jadbabaie2015online, duchi2010composite}.

\begin{theorem}\label{thm_reg_rel_smooth_rel_strcvx}
Under Assumptions 1 and 3, the dynamic regret of online mirror descent with fixed step size $\eta  = 1 /\beta$ satisfies
\begin{align}
\sum_{t=1}^{T}  \Big(f_t( & x_t)  - f_t(u_t)\Big) \leq    \nonumber\\
& (\beta- \lambda) D_r(u_1, x_0) + (\beta - \lambda) \gamma C_T + 2 M G_T ,  \nonumber
\end{align}
for any feasible sequence $\{u_1, u_2, \ldots, u_T\}$, where $M$ represents the diameter of the feasible set, i.e., $M = \max_{x\in \mathcal{X}} \|x\|$, and $C_T$ and $G_T$ denote the path length and gradient variation, as defined in \eqref{eq_path_length} and \eqref{eq_gradient_var_2}, respectively.
\end{theorem}

The proof of Theorem~\ref{thm_reg_rel_smooth_rel_strcvx} is given in App.~\ref{sec_proof_thm_rel_smooth_rel_strcvx}.

 Theorem~\ref{thm_reg_rel_smooth_rel_strcvx} states that the dynamic regret of online mirror descent is upper bounded by $O(1 + C_T + G_T)$. Together with Theorem~\ref{thm_reg_rel_smooth} this immediately leads to the following result.

\begin{corollary} \label{cor_fin}
Under the conditions stated in Theorem~\ref{thm_reg_rel_smooth_rel_strcvx}, the dynamic regret of online mirror descent has an upper bound of $O(1+C_T + \min(V_T, G_T))$.
\end{corollary}

\textit{Remark 2.}
The regularity measures $V_T$ and $G_T$ represent different aspects of an online learning problem, i.e., variation in the functions and gradients. Each of these quantities can be small in an environment that does not change too fast. The resultant bound of $O(1+ C_T + \min(V_T, G_T))$ combines the advantage of these two regularity measures.

\textit{Remark 3.} When the comparator sequence is fixed over time, i.e., $u_1 = \ldots = u_T = \text{argmin}_{x\in \mathcal{X}} \sum_{t=1}^T f_t(x)$, Theorems~\ref{thm_reg_rel_smooth} and \ref{thm_reg_rel_smooth_rel_strcvx} also bound the static regret. In this case, the term involving the path length $C_T$ disappears, and we obtain static regret bounds of $O( 1 + V_T )$ and $O(1 + \min(V_T, G_T))$.

\section{Numerical Experiments}
In this section, we present numerical examples to demonstrate the performance of online mirror descent on cost functions that arise in practice, which are relatively smooth with respect to a carefully chosen regularization function. 

\begin{figure}[t]
\centering
\includegraphics[scale = 0.42]{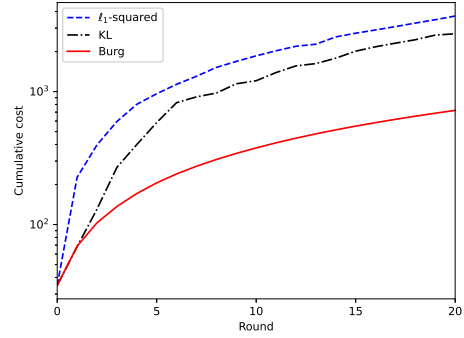}
\caption{Accumulated cost of D-optimal design problem for online mirror descent with the Burg, $\ell_1$-squared, and KL regularization functions.} \label{f_des}
\end{figure}

In the first experiment, we proceed with an application of our algorithmic results to a broad class of D-optimal design problems. The cost functions of interest are $f_t(x) = - \ln \det (H_t \mathbf{D}(x) H_t^T)$, where $\mathbf{D}(x) = \mathrm{diag}(x)$, and $H_t \in \mathbb{R}^{m \times d}$. We not that $f_t(x)$ is neither Lipschitz nor uniformly smooth, but with respect to the Burg regularization function $r(x) = - \sum_{i=1}^{d}\ln (x_i)$ \cite{bauschke2017descent}, it is relatively smooth.
In our experiment, we consider the simplex feasible set, and set $m = 5$ and $d = 10$. In every round $t$, the matrix $H_t$ is selected from a set of randomly generated matrices with independent entries distributed uniformly in $[0, 1]$. 
Fig.~\ref{f_des} shows the accumulated cost of online mirror descent with the Burg regularization function $r(x)$, as well as two of the most commonly used alternatives, namely the $\ell_1$-squared and KL regularization functions. The figure highlights that setting $r(x)$ so that the cost functions enjoy relative smoothness, results in substantially  lower accumulated cost compared with the other common choices of regularization function.

\begin{figure}[t]
\centering
\includegraphics[scale = 0.42]{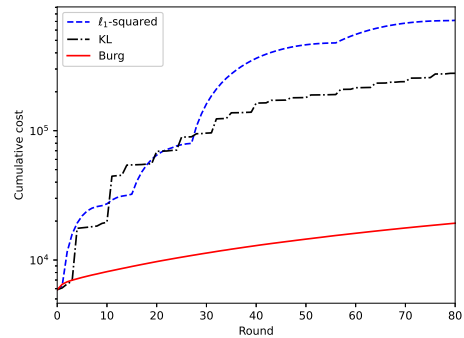}
\caption{Accumulated cost of Poisson linear inverse problem for online mirror descent with the Burg, $\ell_1$-squared, and KL regularization functions.} \label{f_PL}
\end{figure}

Next, we study the application of our results to the Poisson linear inverse problem, which arises in the domain of image science. The cost functions are $f_t(x) = \sum_{i=1}^{m} \Big( b_{t,i} \log ( \frac{b_i}{(A_tx)_i}) + (A_t x)_i - b_{t,i} \Big)$, where $A_t \in \mathbb{R}^{m \times d}_+$ models some experimental protocol, and $b_t \in \mathbb{R}^m_{++}$ is the vector of measurements at round $t$. The goal is to reconstruct the signal $x \in \mathbb{R}^d_{+}$ from the measurements $b_t$ such that $A_t x \simeq b_t$.  We randomly generate $A_t$ and $b_t$ from uniform distribution in $[0,1]$  and set $m = 1500$ and $d = 10$. Since the gradient of $f_t(x)$ is in the order $O(1/x)$, its norm cannot be bounded by a constant, so $f_t(x)$ is neither Lipschitz nor uniformly smooth. However, it can be shown that it is relatively smooth with respect to the Burg regularization function.
Fig.~\ref{f_PL} shows the accumulated cost versus the number of rounds for online mirror descent with the Burg, $\ell_1$-squared, and KL regularization functions. We again observe that Burg regularization performs significantly better than the two commonly used alternatives. 

\section{Conclusion}

In this letter, we provide a new analysis on the  dynamic regret of online convex optimization with mirror descent under relative smoothness. The cost functions do not need to be Lipschitz continuous or uniformly smooth. When the cost functions are relatively smooth, we show that the dynamic regret is bounded by $O(1 + C_T + V_T)$, which depends only on the path length $C_T$ and functional variation $V_T$. In addition, when the cost functions are also relatively strongly convex, we show that the dynamic regret bound can be tightened to $O(1+C_T+ \min(V_T, G_T))$. 
A main observation in these results is that the Hessians of the cost functions can be upper bounded by the Hessian of a carefully designed regularization function.
Our numerical experiments show significant gain in the performance of online mirror descent after choosing an appropriate regularization function with respect to which the cost functions are relatively smooth.


\section*{APPENDIX}

\subsection{Helper Lemmas}

\begin{lemma}\label{lem_ThreePoint} 
For any $x,y,z \in \mathcal{X}$,
\begin{align}  \nonumber
    \langle \nabla r(z) - \nabla r(y), x - y \rangle = D_r(x, y) - D_r(x, z) + D_r(y, z).
\end{align}
\end{lemma} 
The proof of Lemma~\ref{lem_ThreePoint} is given in earlier studies \cite{duchi2010composite, beck2003mirror}.

\begin{lemma}\label{lem_ext_threePoint} 
Let the function $f(\cdot)$ be smooth relative to the function $r(\cdot)$. Then, the following inequality holds for any $x,y,z \in \mathcal{X}$:
\begin{align}  \nonumber
    f(x) - f(y) \leq \langle \nabla f(z), x - y \rangle + \beta D_r(x, z).
\end{align}
\end{lemma} 
The proof of Lemma~\ref{lem_ext_threePoint} is given in \cite{bauschke2019linear}.

\begin{lemma}\label{lem_aux_02} 
Let $\mathcal{X}$ be a convex set and $r(\cdot)$ be a $1$-strongly convex function on $\mathcal{X}$ with respect to some norm. Then, any update of the form
\begin{align}  \nonumber
      x^* = \underset{x \in \mathcal{X}}{\text{argmin}}\Big\{\langle a, x \rangle +  D_r(x, c) \Big\},
\end{align}
satisfies the following inequality:
\begin{align}  \nonumber
      \langle x^* - u, a\rangle \leq D_r(u, c) - D_r(u, x^*) - D_r(x^*, c), \forall u \in \mathcal{X}. 
\end{align}
\end{lemma} 
The proof of Lemma~\ref{lem_aux_02} is given in \cite{shahrampour2017distributed}.

\subsection{Proof of Theorem~\ref{thm_reg_rel_smooth}} \label{sec_proof_thm_rel_smooth}

By the optimality condition of the update \eqref{def_omd}, for any $u \in \mathcal{X}$ we have
\begin{align}
  0 &\leq  \langle \eta \nabla f_t(x_t) + \nabla r(x_{t+1}) - \nabla r(x_t), u - x_{t+1}\rangle \nonumber \\
    & =   D_r(u, x_t) - D_r(u, x_{t+1}) - D_r(x_{t+1}, x_t)   \nonumber\\
    & +  \eta \langle \nabla f_t(x_t), u - x_{t+1} \rangle, \label{q_t1_1}
\end{align}
where the equality follows from Lemma~\ref{lem_ThreePoint}. On the other hand, by applying Lemma~\ref{lem_ext_threePoint} we obtain
\begin{align}
    \eta \Big(   f_t(x_{t+1}) - f_t( & u)  \Big) \leq \label{q_t1_2} \\
    &\eta \Big( \langle \nabla f_t(x_t), x_{t+1} - u \rangle + \beta D_r(x_{t+1}, x_t)  \Big).  \nonumber
\end{align}
By combining \eqref{q_t1_1} and \eqref{q_t1_2}, we have
\begin{align}
    \eta \Big( f_t(x_{t+1} &  )  - f_t(u)  \Big) \leq \label{q_t1_3}  \\
    & D_r(u, x_t) - D_r(u, x_{t+1}) - (1 - \beta \eta) D_r(x_{t+1}, x_t). \nonumber
\end{align}
We then set $\eta = 1/\beta$, so that the last term in the above inequality disappears. We add $f_t(x_t)$ to both sides of \eqref{q_t1_3} and set $u = u_t$ to obtain
\begin{align}
    f_t(x_t) - f_t(u_t) & \leq \frac{1}{\eta} \Big( D_r(u_t, x_t) - D_r(u_t, x_{t+1})  \Big) \nonumber\\
    & + f_t(x_t) -  f_t(x_{t+1}). \label{q_t1_4}
\end{align}
Then, we sum \eqref{q_t1_4} over time to obtain
\begin{align}
\sum_{t=1}^{T} \Big(f_t(x_t) - f_t(u_t)\Big) &\leq \sum_{t=1}^{T} \bigg( \frac{D_r(u_t, x_t)}{\eta} - \frac{D_r(u_t, x_{t+1})}{\eta}  \bigg) \nonumber\\
    & + \sum_{t=1}^{T} \Big( f_t(x_t) - f_{t}(x_{t+1})  \Big). \label{q_t1_5}
\end{align}
We now separately bound the terms on the right-hand side of \eqref{q_t1_5}. In order to bound the first term of the above inequality, we add and subtract several terms as follows:
\begin{align}
    \sum_{t=1}^{T} \bigg( \frac{D_r(u_t, x_t)}{\eta}& -  \frac{D_r(u_t, x_{t+1})}{\eta}  \bigg) = \nonumber \\
    & \sum_{t=1}^{T} \bigg( \frac{D_r(u_t, x_t)}{\eta} - \frac{D_r(u_{t+1}, x_{t+1})}{\eta}  \bigg) \nonumber \\
    & + \sum_{t=1}^{T} \bigg( \frac{D_r(u_{t+1}, x_{t+1})}{\eta} - \frac{D_r(u_t, x_{t+1})}{\eta}  \bigg) \nonumber\\
    & \leq \frac{D_r(u_1, x_1)}{\eta} + \sum_{t=1}^{T} \frac{\gamma\| u_{t+1} - u_t\|}{\eta}, \label{q_t1_6} \raisetag{0.8 cm}
\end{align}
where the last line follows from the fact that $D_r(x,y)$ is non-negative when $r(x)$ is convex, and the Lipschitz condition stated in \eqref{eq_bregman_lip}. 

We now proceed to bound the other term on the right-hand side of \eqref{q_t1_5}. We add and subtract $f_{t+1}(x_{t+1})$ to obtain
\begin{align}
    &\sum_{t=1}^{T} \Big( f_t(x_t) - f_{t}(x_{t+1})  \Big)  = \nonumber\\
    &\sum_{t=1}^{T} \Big( f_t(x_t) - f_{t+1}(x_{t+1}) \Big) + \sum_{t=1}^{T} \Big( f_{t+1}(x_{t+1}) - f_{t}(x_{t+1}) \Big) \nonumber\\
    &\leq f_1(x_1) - f_{T+1}(x_{T+1}) + \sum_{t=1}^{T} \sup_{x \in \mathcal{X}} | f_{t+1}(x) - f_t(x) |. \label{q_t1_7} \raisetag{0.7 cm}
\end{align}
Substituting \eqref{q_t1_6} and \eqref{q_t1_7} into \eqref{q_t1_5} completes the proof.
\hfill $\square$


\subsection{Proof of Theorem~\ref{thm_reg_rel_smooth_rel_strcvx}} \label{sec_proof_thm_rel_smooth_rel_strcvx}

The smoothness of $f_t(\cdot)$ relative to $r(\cdot)$ implies
\begin{align}
    f_t(x_t) &\leq f_t( x_{t-1} ) + \langle \nabla f_{t}(x_{t-1}), x_t - x_{t-1}  \rangle  + \beta D_r(x_t, x_{t-1}) \nonumber\\
             &\leq  f_t( x_{t-1} ) + \langle \nabla f_{t-1}(x_{t-1}), x_t - x_{t-1}  \rangle \nonumber\\
             &+ \langle \nabla f_{t}(x_{t-1}) - \nabla f_{t-1}(x_{t-1}), x_t - x_{t-1}  \rangle \nonumber\\
             &+ \beta D_r(x_t, x_{t-1}). \label{q_t2_1}
\end{align}
To bound the second term on the right hand-side of \eqref{q_t2_1}, we have
\begin{align}
     &\eta \langle \nabla f_{t-1}(  x_{t-1}), x_t - x_{t-1} \rangle =   \nonumber\\
     & \eta \langle \nabla f_{t-1}(  x_{t-1}), u - x_{t-1} \rangle + \eta \langle \nabla f_{t-1}(  x_{t-1}), x_t - u \rangle \leq \nonumber\\
   &\eta \langle \nabla f_{t-1}(  x_{t-1}), u - x_{t-1} \rangle + D_r(u, x_{t-1}) - D_r(u, x_t) \nonumber \\
   & \quad\quad\quad \quad\quad\quad\quad\quad\quad \quad \quad - D_r(x_t, x_{t-1}), \forall u \in \mathcal{X}.    \label{q_t2_2}
\end{align}
where the last line follows from Lemma~\ref{lem_aux_02}.
By combining \eqref{q_t2_1} and \eqref{q_t2_2}, and setting $u = u_t$ we obtain
\begin{align}
    f_t(x_t) &\leq f_{t}(x_{t-1}) + \langle \nabla f_{t}(x_{t-1}) - \nabla f_{t-1}(x_{t-1}), x_t - x_{t-1}  \rangle \nonumber\\
    &+ \langle \nabla f_{t-1}(x_{t-1}), u_t - x_{t-1} \rangle + \frac{D_r(u_t, x_{t-1})}{\eta} \nonumber\\
    &- \frac{D_r(u_t, x_{t})}{\eta} + \Big(\beta - \frac{1}{\eta}\Big) D_r(x_t, x_{t-1}).   \label{q_t2_3}
\end{align}
Furthermore, the strong convexity of $f_t(\cdot)$ relative to $r(\cdot)$ implies
\begin{align}
    f_t(x_{t-1}) + \langle \nabla f_t(x_{t-1}), u_t - x_{t-1} \rangle + \lambda D_r(u_t, x_{t-1}) \leq f_t(u_t).  \label{q_t2_4} \raisetag{0.8cm}
\end{align}
By adding the last inequality to \eqref{q_t2_3}, and setting $\eta = \frac{1}{\beta}$ we have
\begin{align}
f_t(x_t) &\leq f_t(u_t) + \langle \nabla f_{t}(x_{t-1}) - \nabla f_{t-1}(x_{t-1}), x_t - x_{t-1}  \rangle \nonumber\\
& + (\beta - \lambda) D_r(u_t, x_{t-1}) - \beta D_r(u_t, x_t) \nonumber\\
&\leq f_t(u_t) + \| \nabla f_{t}(x_{t-1}) - \nabla f_{t-1}(x_{t-1})\|_*  \|x_t - x_{t-1}\| \nonumber\\
&+ (\beta - \lambda) D_r(u_t, x_{t-1}) -  (\beta - \lambda) D_r(u_t, x_{t}),  \label{q_t2_5}
\end{align}
where the second inequality follows from the fact that $\lambda D_r(u_t, x_t) \geq 0$. We now re-arrange the terms on \eqref{q_t2_5} and sum over time to obtain  
\begin{align}
&\sum_{t=1}^{T}  \Big( f_t(x_t) - f_t(u_t) \Big)\nonumber\\
&  \leq  (\beta - \lambda) \sum_{t=1}^{T} \Big( D_r(u_t, x_{t-1}) - D_r(u_t, x_{t}) \Big) \nonumber\\
& + \sum_{t=1}^{T} \|x_t - x_{t-1}\| \| \nabla f_{t}(x_{t-1}) - \nabla f_{t-1}(x_{t-1})\|_* \nonumber\\
& \leq (\beta- \lambda) D_r(u_1, x_0) \nonumber\\ 
& + (\beta - \lambda) \sum_{t=1}^{T}   \Big(D_r(u_{t+1}, x_{t}) - D_r(u_t, x_{t})\Big) \nonumber\\
&+ \sum_{t=1}^{T} \|x_t - x_{t-1}\| \| \nabla f_{t}(x_{t-1}) - \nabla f_{t-1}(x_{t-1})\|_* \nonumber \\
& \leq (\beta-\lambda) D_r(u_1, x_0) + (\beta - \lambda) \sum_{t=1}^{T}  \gamma \|u_{t+1} - u_t\| \nonumber \\ 
&+ 2M \sum_{t=1}^{T} \sup_{x \in \mathcal{X}} \| \nabla f_{t}(x) - \nabla f_{t-1}(x)\|_*,   \label{q_t2_6}
\end{align}    
where the last inequality follows from the Bregman Lipschitz condition in \eqref{eq_bregman_lip}.
\hfill $\square$

\vspace{-.3cm}
\bibliographystyle{ieeetr}
\bibliography{refs}

\end{document}